\documentclass[lettersize,journal]{IEEEtran}

\usepackage{balance}
\usepackage{times}
\usepackage{multicol}
\usepackage[bookmarks=true]{hyperref}
\usepackage{xcolor}
\usepackage{hyperref}
\usepackage{amsmath, amssymb}
\usepackage{amsfonts}
\usepackage{graphicx}
\usepackage{siunitx}
\usepackage{standalone}
\usepackage{booktabs}
\usepackage[ruled,vlined,linesnumbered, noend]{algorithm2e}
\usepackage{mdframed}
\usepackage{fancyvrb,multirow}
\usepackage{soul}
\usepackage{dsfont,mathabx}
\usepackage{array, booktabs}
\usepackage{subfigure}
\usepackage{amsthm}
\usepackage{makecell}
\usepackage{algorithmic}
\usepackage{etoolbox}

\newtheorem{theorem}{Theorem}
\newtheorem{corollary}{Corollary}

\theoremstyle{definition}
\newtheorem{definition}{Definition}
\newtheorem{remark}{Remark}

\newtheorem{problem}{Problem}

\newcommand{\change}[1]{\textcolor{black}{#1}}

\title{
Accelerated K-Serial Stable Coalition for \\Dynamic Capture and Resource Defense
}

\author{Junfeng Chen, Zili Tang and Meng Guo
\thanks{The authors are with the College of Engineering, Peking University, Beijing 100871, China.
 This work was supported by the National Natural Science Foundation
    of China (NSFC) under grants 62203017, T2121002, U2241214;
    and by the Fundamental Research Funds for the central universities.
  Contact: \texttt{meng.guo@pku.edu.cn}.}
}

\begin{document}

\maketitle
\vspace*{-5mm}


\begin{abstract}
Coalition is an important mean of multi-robot systems to collaborate on common tasks.
An adaptive coalition strategy is essential for the online performance in dynamic and unknown environments.
In this work, the problem of territory defense by large-scale heterogeneous robotic teams is considered.
The tasks include exploration, capture of dynamic targets, and perimeter defense over valuable resources.
Since each robot can choose among many tasks,
it remains a challenging problem to coordinate jointly these robots such that the overall utility is maximized.
This work proposes a generic coalition strategy called K-serial stable coalition algorithm.
Different from centralized approaches,
it is distributed and complete, meaning that only local communication is required
and a K-serial Stable solution is ensured.
Furthermore, to accelerate adaptation to dynamic targets and resource distribution
that are only perceived online,
\change{a heterogeneous graph attention network based heuristic is learned to select
more appropriate parameters and promising initial solutions during local optimization}.
\change{Compared with manual heuristics or end-to-end predictors,
it is shown to both improve online adaptability and retain the quality guarantee.}
The proposed methods are validated via large-scale simulations
with $170$ robots and hardware experiments of $13$ robots,
against several strong baselines such as GreedyNE and FastMaxSum.
\end{abstract}


\section{Introduction}\label{sec:introduction}

Multi-robot systems can be extremely efficient when solving a team-wide task in a concurrent manner.
The key to efficiency is an effective strategy for collaboration to improve overall utility,
where different robots form coalitions to tackle a common task, e.g.,
multiple UAVs surveil an area for intruders by dividing
it into respective regions~\cite{khan2016cooperative,de2019multi};
multiple UGVs capture an intruder by approaching from different
directions~\cite{chung2011search,weintraub2020introduction}.
However, when there are a large amount of diverse tasks to be accomplished,
the optimal collaboration strategy is significantly more difficult to derive,
as each robot needs to choose among numerous potential coalitions,
yielding a combinatorial explosion in complexity~\cite{michalak2016hybrid, rahwan2015coalition}.
\change{Such difficulty is further amplified when the tasks is generated online
non-deterministically and dynamically.}
In these scenarios, an adaptive collaboration strategy is essential to dynamically
adjust coalitions online and re-assign robots to new tasks~\cite{li2020anytime}.
Furthermore, when the numbers of robots and tasks are large,
a centralized coordinator that assigns coalitions centrally quickly
becomes inadequate, especially for the dynamic scenario.
Instead, a decentralized coalition strategy is more
suitable~\cite{delle2012deploying, prantare2020anytime},
where each robot coordinates with other robots via
local communication.

\begin{figure}[t!]
      \centering
      \includegraphics[width=0.95\hsize]{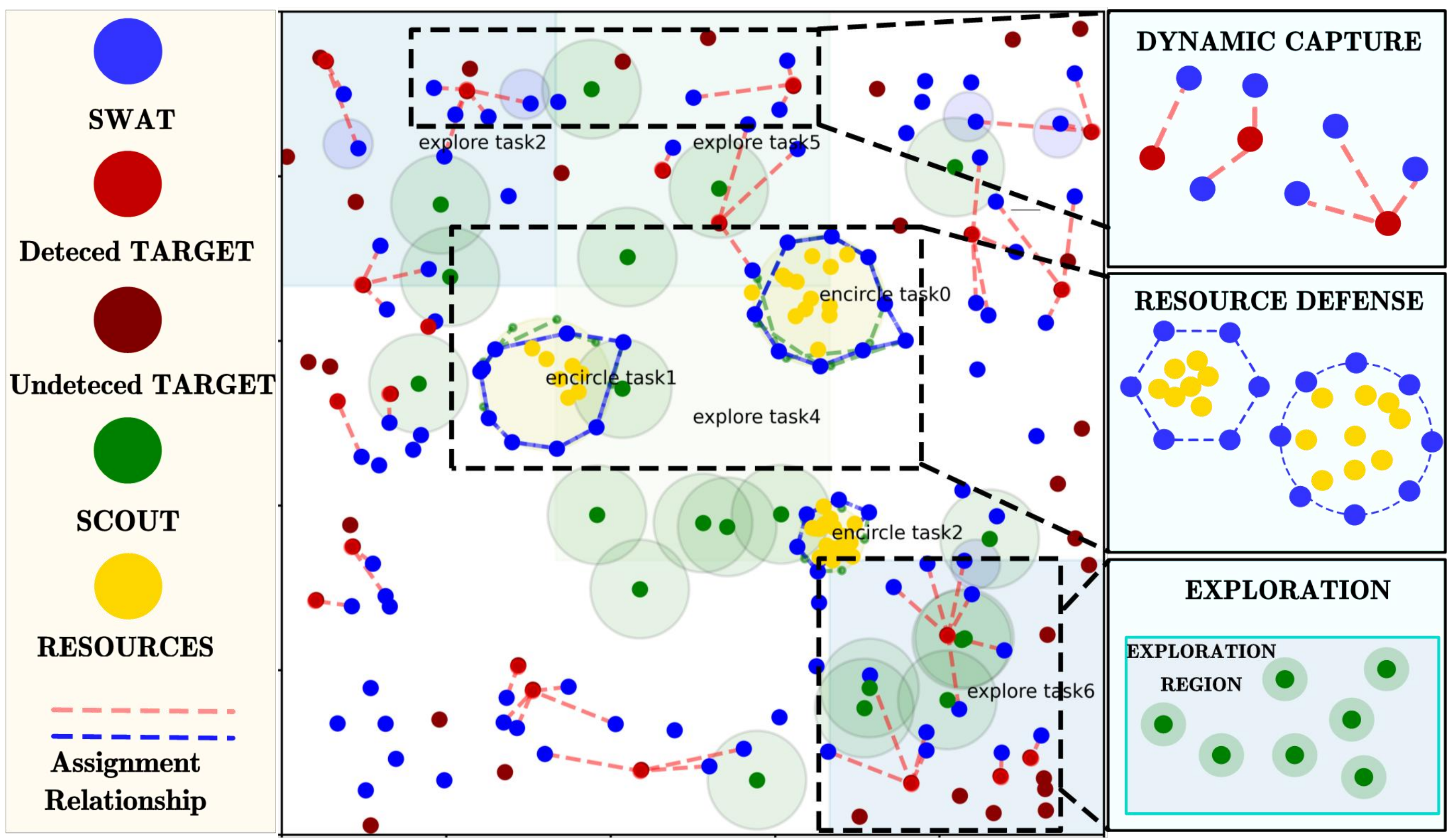}
      \vspace{-2mm}
      \caption{Snapshots of the simulated scenario:
        $100$ UGVs and~$20$ UAVs
        coordinate distributedly and simultaneously three different tasks including
        exploration of over $50$ unknown targets and $100$ unknown resources,
        dynamic capture of detected targets,
        and defense over detected resources.}
      \label{fig:experiment scenario}
      \vspace{-4mm}
\end{figure}

\subsection{Related Work}
This work focuses on the particular application of deploying fleets of UAVs and UGVs
for three different tasks:
the exploration over a territory, the capture of potential intruders,
and the defense of valuable resources.
Much recent work tackles each of these tasks, e.g.,
the collaborative exploration or coverage problem~\cite{khan2016cooperative,de2019multi};
the well-known pursuit-evasion
problem~\cite{chung2011search,weintraub2020introduction,olsen2021visibility};
the perimeter defense problem~\cite{paulos2019decentralization}.
Various extensions can be found where the team size, capability, environment model changes.
For instances, an optimal pursuit strategy is designed in~\cite{garcia2020multiple} first
for the 2v2 perimeter defense problem, and then extended to NvM cases in a greedy way.
The work in~\cite{lopez2019solutions} employs a graph-theoretic approach to
ensure the distance-based observation graph remains connected, i.e.,
the evaders are always visible.
A flexible strategy is proposed in the recent work~\cite{vishnu2021multi},
which can enclose and herd evaders to designated areas.
The work in~\cite{fang2020cooperative} proposes an encirclement
strategy for the slow pursuers to capture the faster evader.
However, these tasks are often considered \emph{separately} and \emph{individually}
in the aforementioned work.
Thus, the proposed methods therein can not be applied directly when the same
team of heterogeneous robots are responsible for a dynamic combination of these three tasks.

On the other hand,
the task assignment for multi-agent systems
refers to the process of assigning a set of tasks to the agents,
see~\cite{torreno2017cooperative,gini2017multi} for comprehensive surveys.
Well-known problems include
the classic one-to-one assignment problem~\cite{jonker1986improving},
the multi-vehicle routing problem~\cite{gini2017multi},
and the coalition structure generation (CSG) problem~\cite{massin2017coalition}.
Representative methods include the Hungarian method~\cite{jonker1986improving},
the mixed integer linear programming (ILP)~\cite{torreno2017cooperative}
and the search-based methods~\cite{fukasawa2006robust}.
However, these methods typically rely on the static table of {agent-task} pairwise cost.
Such cost matrix \emph{does not exist} for any of three tasks considered here.
More specifically, the benefit of one robot joining a coalition depends mainly on
other robots that are already in the coalition,
which often decreases drastically as more robots join the same coalition.
Consequently, these existing methods for task assignment are not suitable
for this application.

Lastly, the CSG problem mentioned above is the most relevant as
it is particularly suitable for collaborative tasks,
see the comprehensive study in~\cite{rahwan2015coalition}.
A marketed-based coalition formation algorithm is presented in~\cite{vig2006multi},
with no clear guarantee on the solution quality.
Provably optimal algorithms are proposed in~\cite{michalak2016hybrid},
which have not been applied to large-scale systems due to combinatorial complexity.
Thus, distributed and anytime approaches are proposed
in~\cite{delle2012deploying, prantare2020anytime},
which are designed to solve static coalition formation problems.
Notably, learning-based methods are proposed in~\cite{de2021decentralized,zhou2022graph}
to learn via reinforcement learning or supervised learning to mimic an expert solver.
\change{However, the learned heuristic in this work is used to \emph{accelerate} the model-based planning algorithm,
  instead of replacing it as a black-box predictor.
  In other words, the theoretical guarantee on solution quality is retained. 
}

\subsection{Our Method}
To overcome these limitations, this work proposes a distributed coalition strategy
for heterogeneous robots under the tasks of dynamic capture and resource defense.
Its backbone is the K-serial stable coalition algorithm (KS-COAL),
which is a distributed and complete algorithm designed for generic definition
of collaborative tasks.
It is particularly suitable for large-scale robotic teams under unknown
and dynamic environments, where the tasks are generated and changed online.
Model-based motion strategies and utility functions are designed for generic tasks,
which are adaptive to team sizes and environment features.
It is proven formally to be complete and strictly better than
the popular Nash-one-stable solutions.
To further improve the solution quality within a limited computation time,
the KS-COAL algorithm is accelerated by choosing more promising initial solutions
and better parameters,
which are learned offline based on optimal solutions of small-scale
problems and use heterogeneous graph attention networks (HGAN) as the network structure.
Consequently, the learned heuristics are integrated seamlessly with the KS-COAL algorithm.
The proposed method is validated extensively on large-scale simulations against
strong baselines.

Main contributions of this paper are summarized as follows:
(i) an unified framework for distributed task coordination of multi-agent systems
under diverse tasks;
(ii) a coordination algorithm with provable guarantee of quality;
(iii) a supervised learning mechanism to accelerate the online coordination process,
without losing the quality guarantee.

\section{Problem Description}\label{sec:problem}

\subsection{Workspace and Robot Description}\label{subsec:ws}
Consider a group of robots~$\mathcal{R}$ that co-exist in a common
workspace~$\mathcal{W}\subset \mathbb{R}^2$.
Each robot~$i\in \mathcal{R}$ is described by three components:
(i) Motion model.
Each robot follows the unicycle model with
state~$s_i$ as the 2D coordinates and orientation,
and input~$u_i$ as the angular and linear velocities;
(ii) Perception model.
Each robot can observe fully the environment within its sensing radius~$r_i>0$;
(iii) Action model.
There are three teams of robots with distinctive capabilities:
(i) SCOUT team~$\mathcal{R}_{\texttt{o}}\subset \mathcal{R}$ that moves swiftly
and is capable of long-range perception;
(ii) SWAT team~$\mathcal{R}_{\texttt{c}}\subset \mathcal{R}$ that is capable of
short-range perception but specialized for capturing other robots
that are less than~$d_{\texttt{c}}>0$ in distance;
and (iii) TARGET team~$\mathcal{R}_{\texttt{s}}\subset \mathcal{R}$ that
is capable of mid-range perception and can take ``resources'' within
the workspace that are less than~$d_{\texttt{s}}>0$ in distance.
In other words, the scout robots, typically UAVs, can navigate easily around
the workspace to locate any target robot and resources,
such that the swat robots, typically UGVs, can approach and capture the
target robots or form encirclement to defend resources.

Lastly, it is assumed that robots within the scout team
can communicate freely without any range limitation,
the same applies to robots within the target team.
However, they can not eavesdrop over other teams communication.
\textcolor{blue}{
All swat robots communicate locally within a limited range,
and the overall communication network is connected
infinitely often.
}

\subsection{Resource Generation}\label{subsec:resources}
To mimic the task of perimeter defense with valuable resources,
it is assumed that one resource~$\xi_t$ is generated every period~$T_{r}>0$ around
several clusters within the workspace,
following a Gaussian Mixture Model (GMM)
e.g.,
$\xi_t\sim \texttt{GMM}(\{(\alpha_k,\mu_k,\Sigma_k)\})$,
where~$(\alpha_k,\mu_k,\Sigma_k)$ are the weight, mean and covariance of Gaussian component~$G_k$.
The distribution of these clusters is static but \emph{unknown} initially, thus can only
be estimated online.
After generation, one resource remains available until it is taken by a target robot.
New resources are constantly generated by the distribution above.
Denoted by~$\Xi_t$ the set of resources that are available at time~$t\geq 0$,
which is set to $\emptyset$ initially.
\change{Meanwhile, each target robot~$i\in \mathcal{R}_\texttt{s}$ searches for resources
via random exploration.
Once found, the robot approaches the resource via the shortest path
and takes it if available, while avoiding being captured by the swat team.
If a target robot is captured, it is immobilized and can not take resources
anymore.}


\subsection{Problem Formulation}\label{subsec:problem-def}
The overall objective is to design a \emph{distributed} control strategy
for the scout and swat teams, such that the number of captured target robots
subtracting the number of resources taken by the target team is maximized.

\begin{remark}\label{remark:distributed}
The desired strategy is distributed,
meaning that there exist {no} central coordinator that assigns globally the
next control and action for each robot.
Instead, the scout and swat robots synthesize the best strategy for the team collaboratively,
via local communication and coordination.\hfill $\blacksquare$
\end{remark}

\section{Proposed Solution}\label{sec:solution}
The proposed solution is a distributed framework for task coordination and motion control.
It consists of three main components:
(i) the motion strategies for all robots are introduced in Sec.~\ref{subsec:motion};
(ii) the distributed coordination algorithm KS-COAL is described in Sec.~\ref{subsec:task};
(iii) a HGAN-based accelerator for the distributed algorithm is designed in Sec.~\ref{subsec:neural}.

\subsection{Task Definition and Motion Strategy}\label{subsec:motion}

\begin{figure}
  \centering
  \includegraphics[scale=0.25]{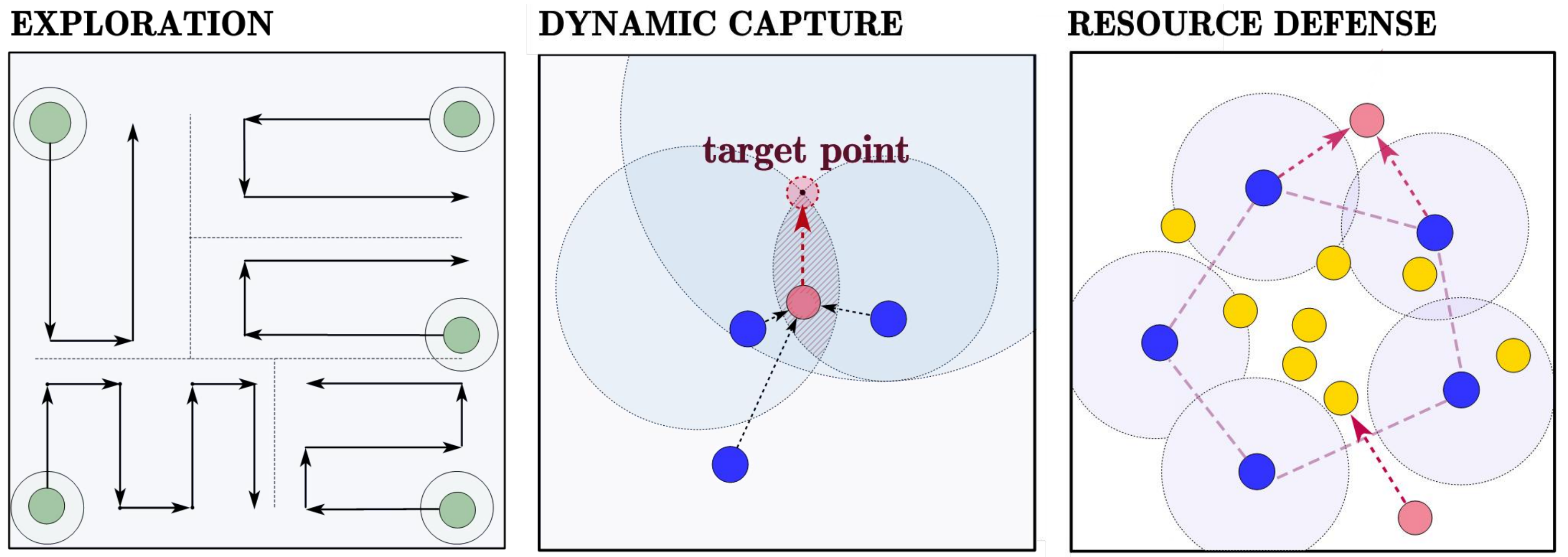}
  \caption{Illustration of the motion strategies for three different tasks.}
  \label{fig:encircle}
  \vspace{-7mm}
\end{figure}

This section describes three different types of tasks and the associated motion strategies
as illustrated as shown in Fig.~\ref{fig:encircle}.
Particularly, the online detection of target robots and resources by the scout robots
is achieved by partitioning the workspace and solving a minimum-time coverage problem~\cite{khan2016cooperative,de2019multi}.
Capture of target robots by the swat robots follows
the dynamic strategy of computing online the intersection of Apollonius circles~\cite{chung2011search,weintraub2020introduction}.
Encirclement of resources by the swat robots is solved as formation problem given the distribution of resources~\cite{paulos2019decentralization,fang2020cooperative}.
\change{The target teams follows the strategy as described in Sec.~\ref{subsec:resources}.}
Additionally, a utility function is defined for possible coalitions w.r.t. each task to measure the required time,
which serves as a fundamental component for the coalition formation algorithm later.
\change{Due to limited space,
details of the task definition and associated motion strategies
are omitted and can be found in the supplementary file.}

\subsection{Distributed Coalition Formation}\label{subsec:task}
Given the above three types of tasks and the associated utility function
described in Sec.~\ref{subsec:motion},
the next prominent step is to form coalitions within the team of swat and scout robots
such that the overall utility for the team is optimized.
\change{Different from most relevant work that focuses on a centralized solution
  for static scenes,
this work proposes a \emph{unified}, \emph{complete} and \emph{distributed} algorithm
that can handle a large number of diverse tasks in dynamic environments.}

\subsubsection{Problem Reformulation}\label{subsubsec:prob}
To begin with, the task assignment problem is formulated
based on the coalition formation literature~\cite{rahwan2015coalition,li2020anytime}.
Particularly, consider the following definition of a coalition structure:
\begin{equation}\label{eq:general-form}
 \mathcal{F} \triangleq (\mathcal{R},\,\Omega,\, f,\,\mathcal{K}),
\end{equation}
where~$\mathcal{R}=\{1,\cdots,N\}$ is the team of~$N$ robots;
$\Omega=\{\omega_1,\cdots, \omega_M\}$ is the set of~$M$ tasks;
$f:2^{\mathcal{R}} \times \Omega \rightarrow \mathbb{R}^+$ is the utility function
of a potential coalition for any given task in~$\Omega$;
\change{
$\mathcal{K}=\{k_1,\cdots,k_N\}$ is the optimality index designed for
 limiting the communication and computation complexity for each robot,
which is assumed to be static for now
but adaptive in the overall algorithm presented in the sequel.}
The set of tasks that {can} be performed by robot~$i$ is denoted by~$\Omega_i$.

\begin{definition}[Assignment]\label{def:assignment}
A valid solution of the coalition structure~$\mathcal{F}$ in~\eqref{eq:general-form} is called an assignment,
denoted by~$\nu =\{(\omega_m,\,\mathcal{R}_m),\,\forall \omega_m \in \Omega\}$,
where~$\mathcal{R}_m \subset \mathcal{R}$ represents a coalition
that performs the common task~$\omega_m \in \Omega$,
and~$\mathcal{R}_{m_1}\cap\mathcal{R}_{m_2}=\emptyset,\, \forall m_1\neq m_2$.
Moreover, an assignment is called optimal if its {mean} utility, i.e.,
\begin{equation}\label{eq:cost}
\varrho(\nu) \triangleq \frac{\Sigma_{\omega_m\in \Omega}\,f(\mathcal{R}_m,\, \omega_m)}{M},
\end{equation}
is maximized, and denoted by~$\nu^\star$.
\hfill $\blacksquare$
\end{definition}

\change{Note that the mean utility above is chosen to be consistent with the main objective,
and resembles the overall success rate in dynamic scenes.}
With a slight abuse of notation,
let $\nu(\omega_m)=\mathcal{R}_m$ which returns the coalition for task~$\omega_m$
and conversely, $\nu(i)=\omega_m$ which returns the  task assigned to robot~$i$,
$\forall i \in \mathcal{R}_m$.
More importantly, an assignment~$\nu$ can be modified
via the following {switch} operation by any robot.
\begin{definition}[Switch Operation]\label{def:switch}
  The operation that robot~$i \in\mathcal{R}$ is assigned to
  task~$\omega_{m} \in \Omega$ is called a \emph{switch} operation,
  denoted by~$\xi^{m}_{i}$.
  The switch~$\xi^{m}_{i}$ is valid only if~$\omega_m \in \Omega_i$.
  Thus, via~$\xi^{m}_{i}$, an assignment~$\nu$ is changed into a new assignment~$\hat{\nu}$,
  such that~$\hat{\nu}(i)=\omega_m$.
  For brevity, denote by~$\hat{\nu}=\xi^{m}_{i}(\nu)$.
  \hfill $\blacksquare$
\end{definition}

\begin{definition}[Chain Transformation]\label{def:transformation}
A \emph{transformation} is defined as a chain of allowed switches, i.e.,
\begin{equation}\label{eq:transformation}
\Xi \triangleq \xi_{i_1}^{m_1} \circ \xi_{i_2}^{m_2} \circ \xi_{i_\ell}^{m_\ell} \cdots \circ \xi_{i_L}^{m_L},
\end{equation}
where~$L$ is the total length;
$i_{\ell+1} \in \mathcal{N}_{\ell}$ is the constraint of forming a chain,
$\forall \ell = 0,\cdots, L-1$.
The transformation $\Xi$ is valid if all switches inside are valid.
Thus, via~$\Xi$, an assignment~$\nu$ can be transformed into a new assignment~$\hat{\nu}$,
by recursively applying the switches~$\xi_{i_\ell}^{m_\ell}\in \Xi$,
such that~$\hat{\nu}(i_\ell)=\omega_{m_\ell}$, $\forall \ell = 0,\cdots, L-1$.
For brevity, denote by~$\hat{\nu}=\Xi(\nu)$.
 \hfill $\blacksquare$
\end{definition}

\begin{definition}[Rooted Transformation]\label{def:root-transformation}
  A rooted transformation at robot~$i\in \mathcal{R}$ is denoted by~$\Xi_i$,
  which is a chain transformation that starts from~$i$,
  i.e.,~$i_1=i$ in~\eqref{eq:transformation}.
 \hfill $\blacksquare$
\end{definition}
\begin{remark}
  The chain transformation above is different from ``arbitrary modification'' in \cite{li2020anytime},
  where the sequence of transformations can be operated on any robot.
  However, the chain structure in~\eqref{eq:transformation} requires that only neighboring robots
  can perform switches consecutively.
  The main motivation is to limit the search space of possible transformations to robots within
  the~$K$-hop local communication network,
  thus accelerating the convergence to a high-quality solution.
\hfill $\blacksquare$
\end{remark}


\begin{definition}[$\mathcal{K}$-Serial Stable]\label{def:k-serial}
An assignment $\nu^\star$ is $\mathcal{K}$-Serial stable (KSS),
if for each robot~$i\in \mathcal{R}$,
there does not exist \emph{any} rooted transformation~$\Xi_i$ such that:
(i) $\Xi_i$ is valid;
(ii) $\varrho(\Xi_i(\nu^\star)) > \varrho(\nu^\star)$;
(iii) $|\Xi_i|\leq k_i$,
where~$k_i$ is the optimality index defined in~\eqref{eq:general-form}.
\hfill $\blacksquare$
\end{definition}

\begin{remark}\label{remark:ks}
  The above notion of KSS solution
  contains the Nash equilibrium in~\cite{li2020anytime}
  and the globally optimal solution as special cases.
  Namely, if the indices are chosen that~$k_i=1$, $\forall i\in \mathcal{R}$,
  then the KSS solution is equivalent to the Nash equilibrium
  as at most one robot is allowed to switch its own task.
  If~$k_i=N$, $\forall i\in \mathcal{R}$
  and the communication graph is connected,
  then the KSS solution is equivalent to the globally optimal solution
  as all robots can switch tasks simultaneously.
  Thus, the parameter~$\mathcal{K} = \{k_i\}$
  provides the flexibility to adaptively improve
  the solution given the time budget.
\hfill $\blacksquare$
\end{remark}

\begin{problem}\label{prob: k-serial-stable-solution}
  Given the coalition structure~$\mathcal{F}$ in~\eqref{eq:general-form},
  design a distributed algorithm to find one KSS solution by Def.~\ref{def:k-serial}.
\hfill $\blacksquare$
\end{problem}

\subsubsection{Algorithm Description}\label{subsubsec:alg}
\begin{figure}[!t]
  \centering
  \includegraphics[width=0.9\hsize]{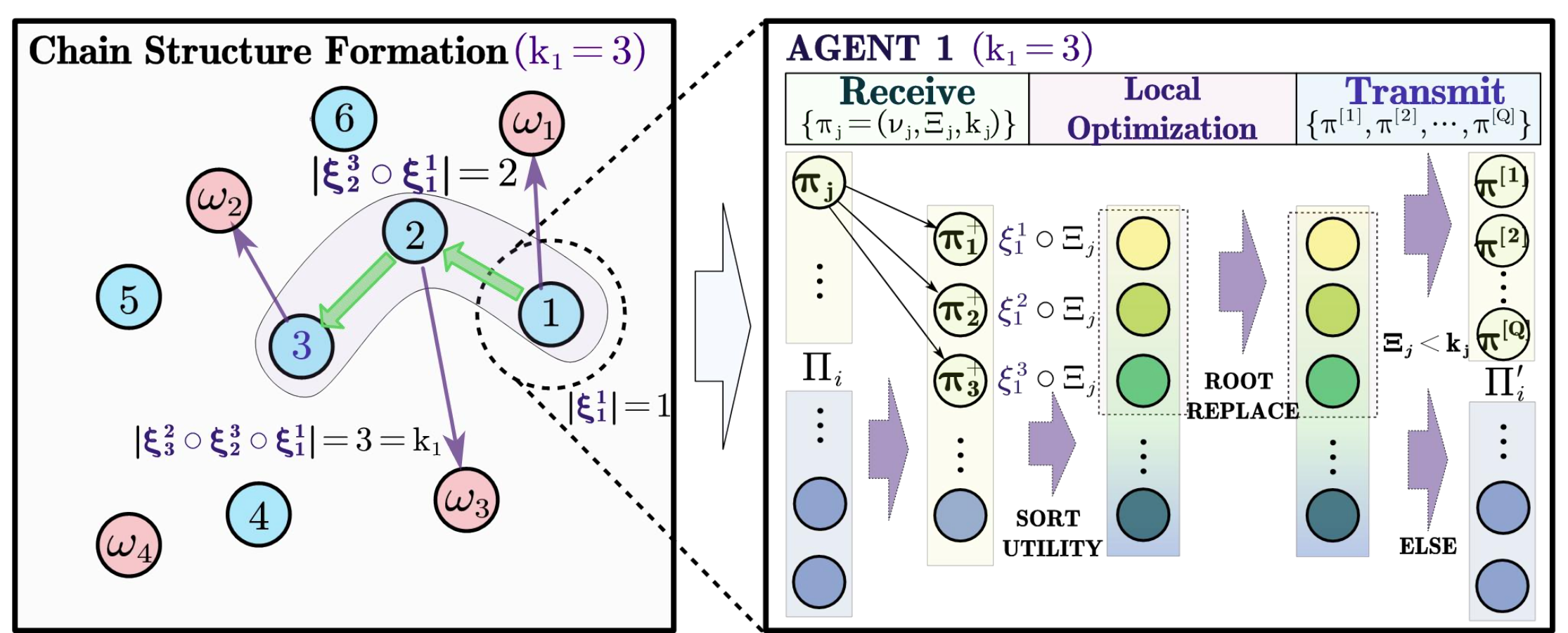}
  \vspace{-2mm}
  \caption{Illustration of key steps in the KS-COAL algorithm:
    distributed exchange via the chain structure (\textbf{left})
    and local optimization (\textbf{right}).}
  \label{fig:ks-coal}
  \vspace{-4mm}
\end{figure}

A distributed, asynchronous and complete algorithm
called the $\mathcal{K}$-serial stable coalition ({KS-COAL})
is proposed to solve Problem~\ref{prob: k-serial-stable-solution}.
\change{As shown in Fig.~\ref{fig:ks-coal}, the main idea is to combine distributed
communication and local optimization.}
Each robot~$i$ attempts to find the locally-best transformation rooted at itself,
which can improve the currently-known best solution the most.
In addition, each robot also participates in the optimization of transformations
rooted at other robots, i.e., as part of the chain defined in~\eqref{eq:transformation}.
\change{Lastly,
a consensus protocol is running in parallel to encourage
the convergence to a common KSS solution across the team.}

More specifically, as summarized in Alg.~\ref{alg:search},
each robot follows the same algorithmic procedure asynchronously.
\change{
  At~$t=0$, each robot computes the set of initial partial solutions
  by assigning one task to itself, as its local information.}
Then, robot~$i$ sends the message below to all its neighbors~$j \in \mathcal{N}_i$:
\begin{equation}\label{eq:msg}
  \pi_{i\rightarrow j}\in \{(\nu_0,\,\varepsilon,\, k_i), \; \nu_0(i)=\omega_m \in \Omega_i\},
\end{equation}
where $\nu_0$ is the partial solution;
$\varepsilon$ denotes the null transformation and~$k_i$ is the optimality index of robot~$i$.
This message is added to a local message buffer denoted by $\Pi_i$, see Line~\ref{alg:init}.

\change{At $t > 0$, each robot~$i\in \mathcal{N}$ follows the same round of
  receiving messages, local optimization, sending messages,
  and consensus, as in Lines~\ref{alg:loop_begin}-\ref{alg:loop_end}.}
To begin with, for any stored message $\pi = (\nu, \, \Xi,\, k) \in \Pi_i$ that $|\Xi|\leq k$ holds,
robot~$i$ relays this message to its neighbors, i.e., $\pi_{i \rightarrow j}=\pi$, $\forall j\in \mathcal{N}_i$.
Meanwhile, after receiving each message~$\pi_{j\rightarrow i}$ from robot~$j$ that~$i\in \mathcal{N}_j$,
it stores~$\pi_{j\rightarrow i}$ in its buffer~$\Pi_i$.
During local optimization,
robot~$i$ generates a new message~$\pi^+$ for any
selected message~$\pi=(\nu,\,\Xi,\,k)\in \Pi_i$
by appending an additional switch operation for \emph{each} task $\omega_m \in \Omega_i$ to the existing transformation, i.e.,
\begin{equation}\label{eq:local-operation}
\pi^+ = (\nu,\, \Xi \circ \xi^m_i,\, k),\, \forall \omega_m \in \Omega_i;
\end{equation}
where~$\nu$ is the same initial solution as in~$\pi$;
$\xi^m_i$ assigns robot~$i$ to task~$\omega_m$.
Note that this modification is only allowed if~$|\Xi|\leq k$,
i.e., its length limit is not reached.
This new message~$\pi^+$ is stored in~$\Pi_i$,
see Line~\ref{alg:switch}-\ref{alg:switch_store}.
More importantly, robot~$i$ maintains the best assignment
that is known to itself, denoted by~$\nu^\star_i$.
It is updated each time a message is received
or a new message is generated in \eqref{eq:local-operation}, i.e.,
\begin{equation}\label{eq:update-nu}
  \nu^\star_i = \Xi(\nu), \quad \text{if}\;\rho(\Xi(\nu)) > \rho(\nu^\star_i),
\end{equation}
as in Line~\ref{alg:update_best},
where the updated utility~$\rho(\Xi(\nu))$ can be computed locally by robot~$i$
as~$\rho(\Xi(\nu))=\rho(\nu^\star_i)+\Delta_f (\omega_m) + \Delta_f (\omega'_m)$,
where~$\omega_m,\,\omega'_m$ are tasks assigned to robot~$i$ before and
after the update, respectively;
and~$\Delta_f(\cdot)$ is the difference in task utility
of~$\omega_m,\,\omega'_m$ before and after the update.

\begin{algorithm}[t]
  \label{alg:search}
  \caption{KS-COAL algorithm.}
  \SetAlgoLined
  \KwIn{Coalition structure~$\mathcal{F}$.}
  \KwOut{KSS solution~$\nu^\star$.}
  \tcc{Message passing}
  Initialize~$\Pi_i$ by~\eqref{eq:msg}, \, $\Pi^\star_i = \emptyset$ \; \label{alg:init}
  \While{not terminated}{
    \tcc{Sending message}
  \For{$\pi=(\nu,\,\Xi,\,k) \in \Pi_i$ and $|\Xi| \le k$}{ \label{alg:loop_begin}
    Send~$\pi_{i \rightarrow j}$ to $\mathcal{N}_i$\;
    Remove $\pi_{i \rightarrow j}$ from $\Pi_i$\;}
  \tcc{Receiving message}
  \ForAll{$i\in \mathcal{R}$}{
      Receive $\{\pi_{j \rightarrow i}\}$ and update~$\Pi_i$\;}
  \tcc{Local optimization}
  \For{$\pi=(\nu,\,\Xi,\,k) \in \Pi_i$}
      {Compute $\xi_{i}^{m}$ and~$\pi^+$ by~\eqref{eq:local-operation}\; \label{alg:switch}
      Save $\pi^+$ to~$\Pi_i$\; \label{alg:switch_store}
      Update~$\nu_i^\star$ by~\eqref{eq:update-nu}\;}  \label{alg:update_best}
  \tcc{Consensus}
  \uIf{$\varrho(\nu_j^\star)>\varrho(\nu_i^\star)$}{ \label{alg:best_computation}
    $\nu_i^\star \leftarrow \nu_j^\star$\;
    Compute~$\{\Xi_i(\nu^\star_i)\}$ and update~$\Pi_i$ by~\eqref{eq:update-nu-star}\; \label{alg:root_replacement}
    Update~$\nu_i^\star$ by~\eqref{eq:update-nu}\; \label{alg:loop_end}
  }
  }
\end{algorithm}
\change{
As mentioned earlier,
a consensus protocol is executed in parallel,
such that the team converges to the same KSS solution.}
However, this should be done in special care to avoid pre-mature termination
and over-constrained search space.
In particular, each robot~$i$ shares its best solution~$\nu^\star_i$
in~\eqref{eq:local-operation} with its neighbors,
i.e., by sending directly the associated message~$\pi^\star_i$,
then robot~$i$ can locally compute the best assignments received from other robots,
namely, $\rho(\nu^\star_j)$ as in Line~\ref{alg:best_computation}.
Afterwards, robot~$i$ updates its best assignment
if any received assignment yields a better overall utility,
i.e., $\rho(\nu^\star_j)>\rho(\nu^\star_i)$.
Then, all rooted transformations~$\{\Xi_i\}$ stored in~$\Pi_i$
should be applied to~$\nu^\star_i$
to generate new solutions~$\{\Xi_i(\nu^\star_i)\}$, i.e.,
\begin{equation}\label{eq:update-nu-star}
(\nu,\,\Xi_i,\,k_i) \rightarrow (\nu^\star_i,\,\Xi_i,\,k_i),
\end{equation}
as in Line~\ref{alg:root_replacement}.
Meanwhile, these invalid messages are removed from $\Pi_i$.
Thus, the new assignment is given by~$\Xi_i(\nu^\star_i)$,
of which the utility is computed by tracing back the chain operations in~$\Xi_i=\xi^{m_1}_{i_1}\circ \xi^{m_2}_{i_2} \cdots \circ \xi^{m_L}_{i_L}$
and querying each robot~$i_\ell$ along the chain directly the change of utility.
If any of these new assignments in~$\{\Xi_i(\nu^\star_i)\}$ yields a
better utility than~$\nu^\star_i$,
it replaces~$\nu^\star_i$ and is shared with the neighbors,
see in Line~\ref{alg:loop_end}.
If the local best assignment~$\nu^\star_i$ remains unchanged
after a certain number of rounds,
the algorithm terminates.
\change{Note that the above algorithm is anytime,
  meaning that it can be interrupted anytime given the limited computation resources
  and still yield the currently-best solution.}

\subsubsection{Performance Analysis}\label{subsubsec:solu_qual}
Under the proposed algorithm, convergence to a KSS solution is ensured.
Furthermore, it is shown that this solution extends naturally to
the globally optimal solution if~$\mathcal{K}$ is set accordingly.
Proofs are omitted here and provided in the supplementary file.

\begin{theorem}\label{theory:stable}
  Under Alg.~\ref{alg:search}, the local assignment is ensured to converge to~$\nu^\star_i = \nu^\star$,
  $\forall i \in \mathcal{R}$, where~$\nu^\star$ is a KSS solution.
\end{theorem}
\begin{corollary}\label{the:K-serial To global optimal}
  Given that~$k_i = N$, $\forall i\in \mathcal{R}$,
  the KSS solution return by Alg.~\ref{alg:search} is
  the globally optimal solution.
\end{corollary}

\subsection{Neural Accelerator for Dynamic Adaptation}\label{subsec:neural}
Since all robots are
\emph{constantly} moving and resources are generated dynamically,
the current assignment might not be optimal or even infeasible at latter time.
In many cases only a few robots need to adjust their assignments.
Thus, a neural accelerator (NAC) based on heterogeneous
graph attention networks (HGAN) from~\cite{wang2019heterogeneous}
is proposed to accelerate the proposed KS-COAL,
by
predicting directly the choice of parameter~$\mathcal{K}$
and an initial solution for KS-COAL.

\subsubsection{Network Structure of NAC}\label{subsubsec:network}
\begin{figure}[t!]
    \centering
    \includegraphics[height=0.4\hsize]{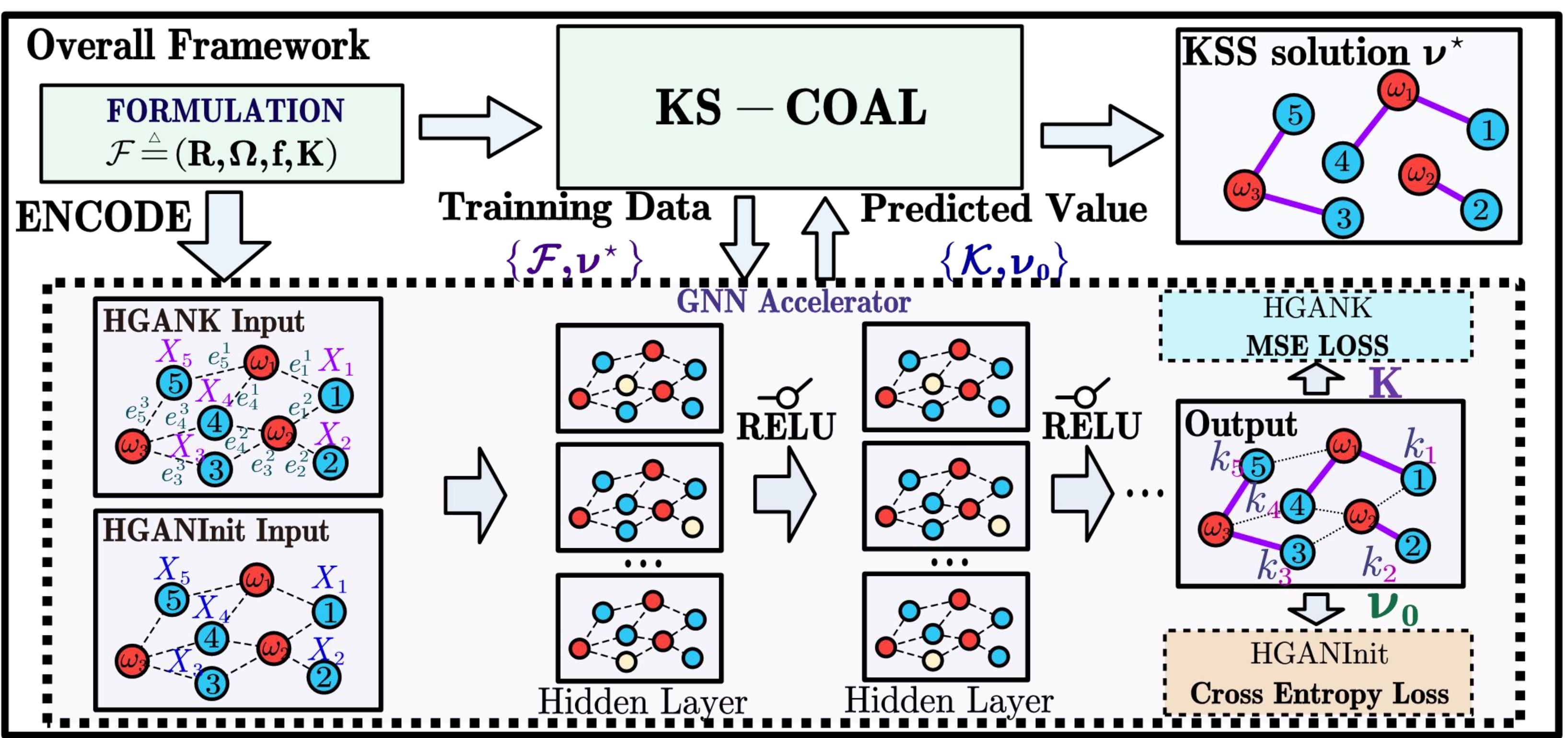}
    \vspace{-2mm}
    \caption{Network structure of the proposed NAC:
      the HGAN-K branch (\textbf{upper}) and the HGAN-Init branch (\textbf{lower})
      that predicts the parameter~$\mathcal{K}$
      and initial solution~$\nu_0$,
      which are used to accelerate the KS-COAL algorithm.}
    \label{fig:gnn}
    \vspace{-5mm}
\end{figure}

As shown in Fig.~\ref{fig:gnn},
the proposed HGAN model contains two branches called HGAN-K and HGAN-Init,
which are used to predict the $\mathcal{K}$ and $\nu_0$, respectively.
They both share the common two HGAN middle-layers to extract the hidden embedding.
As inputs to both branches, two labeled graphs are constructed based on
the same underlying coalition structure.
In particular, three types of robots and three types of tasks
are encoded as vertices with different attributes.
Undirected edges from robots to tasks indicate whether one robot can participate in one task.
Both node and edge attributes are encoded \emph{differently} within HGAN-K and HGAN-Init:
(i)
The attribute of each edge in HGAN-K is a vector of marginal utility given by:
\begin{equation*}\label{eq:margin-utility}
  \upsilon_m(i,\omega_m) = \left\{f(\mathcal{R}_m,\, \omega_m)
  -f(\mathcal{R}_m \backslash \{i\},\, \omega_m),
  \, \forall \mathcal{R}_m \in \widehat{\mathcal{R}}_m\right\},
\end{equation*}
where~$i\in \mathcal{R}_m$ is any coalition that can perform task~$\omega_m\in \Omega$;
$\widehat{\mathcal{R}}_m\subset 2^{\mathcal{R}}$ is the set of all such coalitions;
$\upsilon \in \mathbb{R}^{|\widehat{\mathcal{R}}_m|}_{\geq 0}$ is a vector of non-negative reals.
However, the edges for HGAN-Init are attribute-free and are only responsible for passing messages.
(ii)
Each vertex in HGAN-K is labeled by their robot or task type.
However,
each vertex in HGAN-Init is encoded by stacking~$\upsilon_m(i,\omega_m)$ above for all possible tasks.
Zero-padding is applied to ensure that these attributes have consistent sizes across vertices and edges.
These graphs are then processed by the HGANs tailored for heterogeneous graphs with varying sizes.
The HGAN-K outputs the predicted parameter~$\mathcal{K}$ over each robot vertex,
while HGAN-Init outputs a ranking over all edges of each robot vertex.
More details of the adopted HGAN structure can be found in the supplementary file.

\subsubsection{Training and Execution}\label{subsubsec:train}
To begin with,
the training data is collected by solving the considered scenario by the proposed KS-COAL,
under different random seeds.
More specifically, it is given by~$\mathcal{D}=\{(\mathcal{F},\,\mathcal{K},\,\nu^\star)\}$,
where the tuple consists of a problem instance~$\mathcal{F}$ in~\eqref{eq:general-form}
under a random choice of~$\mathcal{K}$;
$\mathcal{K}$ is the optimal choice for each robot,
and the associated KSS assignments~$\nu^\star$ obtained by KS-COAL.
First, each problem instance is transformed into labeled graphs as described above,
\change{while the optimal robot-task pairs are derived from~$\nu^\star$
and the optimal choice of~$\mathcal{K}$ is quantized into~$\{1,2,3\}$ given~$J^\star$,
where $J^\star\in \mathbb{Z}^N$ is the total number of switches
that each robot~$i\in\mathcal{N}$ has performed \emph{and} resulted
in an increase of the overall utility.}
Then, the HGAN-K and HGAN-Init branches are both trained in a supervised way,
where the ``weighted MSE loss'' is chosen as the loss function for HGAN-K
to avoid imbalanced data,
and the ``Cross Entropy loss'' for HGAN-Init.
\change{Note that during training, the modern HGANs recursively update the node and edge features
by aggregating information from multiple-hop neighbors.}
Consequently, the trained HGAN-K can predict the optimal choice of~$\mathcal{K}$,
while the HGAN-Init outputs directly a recommended solution~$\nu_0$.

During execution, each robot constructs its local multi-hop graphs given a new problem~$\mathcal{F}$,
which is encoded in the same way as the training process.
\change
{Then, this labeled local graph is fed into the learned HGAN to obtain the hidden embedding,
which is then aggregated via local communication with neighboring robots.}
Consequently, after convergence, the aggregated embedding is used to predicate the optimal
choice of~$k_i$ and $\omega_m$ for each robot~$i\in \mathcal{N}$.
Thus, the KS-COAL algorithm can be kick-started with this choice of~$\mathcal{K}$
and this initial assignment~$\nu_0$.
It is worth mentioning that the guarantee on solution quality in Theorem~\ref{theory:stable}
is still retained.

\vspace{-1.5mm}

\subsection{Computation Complexity}\label{subsec:summary}
Since Alg.~\ref{alg:search} is fully distributed,
its complexity is analyzed for a single robot.
For one round of communication,
robot $i$ can send at most $ |\Pi_i| $ messages to $ |\mathcal{N}_i| $ robots.
Thus, the worst time-complexity for sending messages is $\mathcal{O}(N_{\Pi} N_r)$,
where $ N_{\Pi} $ is the upperbound of $ |\Pi_i| $ and
$ N_{r}<N$ is the upperbound of $|\mathcal{N}_i|$.
Similarly, the stage of ``local optimization'' has
the time-complexity~$\mathcal{O}(N_\Pi N_\omega)$,
where $N_\omega$ is the maximum number of the tasks that one robot is capable.
Thus, the overall time-complexity at each iteration is~$\mathcal{O}(N_{\Pi} (2N_{r}+N_\omega))$.
Note that $N_{\Pi}$ is upper bounded by
$\left(N_\omega N_{r}\right)^{k_{\max}}$,
where $k_{\max}=\max_i\, \{k_i\}$.
Furthermore,
since the exact time of convergence~$\bar{t}$ is hard to determine,
a factor $\epsilon$ is introduced to relax the convergence criterion.
More specifically,
it can be regarded as ``convergence'' if there is no $\Xi$ that can
improve the utility~$\rho$ by more than $\epsilon$.
For a single robot, the number of effective switches is upper bounded by~$\Delta \rho /\epsilon$,
where $\Delta \rho$ is the upper bound of  $\rho^\star-\rho$.
Since the iterations required for ``consensus'' is bounded by~$N$ between
two effective switches, it holds that~$\bar{t} \leq \frac{\Delta \rho}{\epsilon} N$.
Thus, the overall time-complexity of Alg.~\ref{alg:search} is given by
$\mathcal{O}((N_\omega N_{r})^{k_{\max}}(2N_r+N_{\omega})(\Delta \rho/\epsilon)N)$.

\begin{figure}[!t]
  \centering
  \includegraphics[width=0.98\hsize]{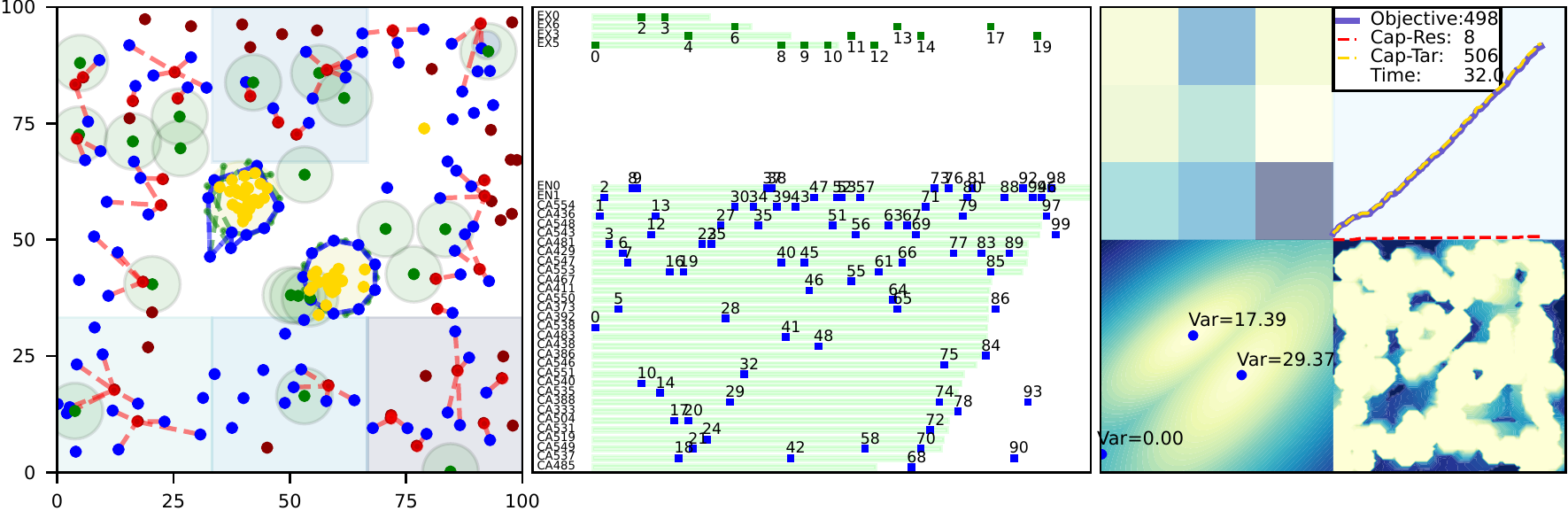}
    \vspace{-2mm}
    \caption{Snapshot of the task execution:
     motion of all robots (\textbf{left}),
     dynamic task assignment (\textbf{middle}),
     estimated distribution of resources and targets (\textbf{bottom-right}),
     number of captured targets and resources (\textbf{top-right}).
    }
  \label{fig:evolution}
  \vspace{-5mm}
\end{figure}

\section{Numerical Experiments} \label{sec:experiments}
Extensive numerical experiments are conducted for large-scale systems,
which are compared against several strong baselines.
The proposed method is implemented in Python3 with Pytorch geometric Graph (PyG)
from~\cite{Fey/Lenssen/2019}
and tested on an AMD 7900X 12-Core CPU @ 4.7GHz with RTX-3070 GPU.
More detailed descriptions and experiment videos can be found in the supplementary file.

\subsection{Setup}

\subsubsection{Agent and Workspace Model}\label{subsec:model}
As shown in Fig.~\ref{fig:evolution},
consider $100$ swat robots, $20$ scout robots, and $50$ target robots in a free workspace of size~$100m \times 100m$.
Their reference velocities are set to $1m/s$, $6m/s$ and $0.9m/s$.
The sensing radius is set to~$5m$ for swat robots and $10m$ for scout robots.
The capture range of swat and target robots are set to $3m$ and $2m$.
Moreover, the resources are generated by three Gaussian clusters of variance~$0.75m$ every~$5s$.
Note that the initial distribution of target robots and resources are unknown to the swat and scout robots.
Captured targets and taken resources are removed from display.
In addition, the workspace is divided into $9$ regions for the exploration task.
\change{The proposed method is triggered
by specific events and at regular intervals,
specifically when targets are captured and every $5$ time steps.}
Detailed descriptions can be found in supplementary file.

\subsubsection{Proposed Method and Baselines}
The detailed setup of the proposed \change{NAC} is omitted here and
provided in supplementary file.
\change{In total~\textbf{10 baselines} are compared:}
(i)
\change{the KS-COAL algorithm with the parameter~$\mathcal{K}$ uniformly set to~$2$, $3$, $4$ or randomly,
denoted by KS2, KS3, KS4 and Random, respectively};
(ii) two variants of the \change{NAC} for ablation study:
\change{NACK} that only uses the learned accelerator to predict~$\mathcal{K}$
with the initial solution being the previous solution,
and \change{NACInit} that only predicts the initial solution with a random $\mathcal{K}$;
(iii) two state-of-the-art methods for distributed coordination:
\change{GreedyNE} from~\cite{prantare2020anytime},
where a Nash-stable assignment is selected;
\change{FastMaxSum} from~\cite{li2022task},
which solves the matching problem over a task-robot bipartite graph via distributed message passing.
(iv)
\change{
  two end-to-end learning paradigms are implemented,
  i.e., reinforcement learning from~\cite{de2021decentralized}
  and supervised learning from \cite{zhou2022graph},
  denoted by RL and HGAN.
}

\subsection{Results and Comparisons}\label{subsec:results}


\subsubsection{Overall Performance}
As illustrated in Fig.~\ref{fig:evolution},
the robot motion and actions are monitored during execution,
along with the dynamic task assignment.
In addition, the detected targets and resources
are shown with the number of captured targets and preserved resources.
\change{Note that each time the proposed NAC is triggered,~$20$ rounds of
  communication are allowed within the team.}
It can be seen that
the scout robots are assigned to different partitions of the workspace.
Three clusters of resources are gradually identified.
Once the swat robots receive the position of target robots from the scout robots,
coalitions are formed dynamically to capture target robots and form
encirclement around the clusters of resources.
Note that some swat robots become immobile
when the marginal utility of joining a capture or defense task is negligible.
Also, swat robots may switch from a defense task to a capture task,
when some target robots are nearby.
In total, $50$ resources are kept and $234$ target robots are captured
over~$200$ time steps.




\begin{figure}[t!]
    \centering
    \includegraphics[width=0.86\hsize]{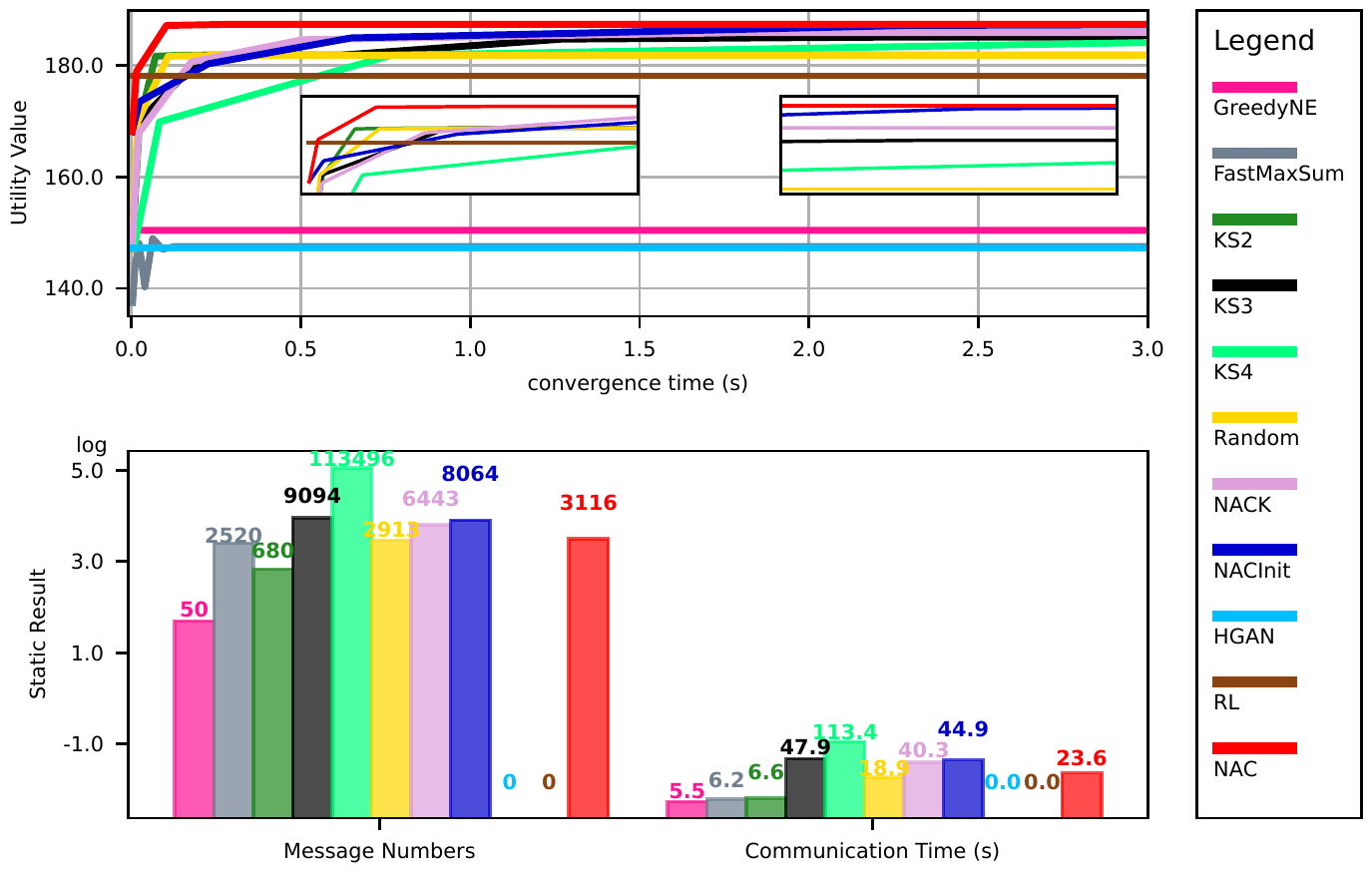}
      \vspace{-4mm}
    \caption{\change{Evolution of the overall utility along
    with the computation time for our method~\textbf{NAC}
    and~$10$ other baselines (\textbf{top});
    the associated total number of messages and the time of communication (\textbf{bottom}).}}
    \label{fig:static_utiltiy}
    \vspace{-4mm}
  \end{figure}

\subsubsection{Convergence over a Fixed Problem}
To begin with, the baselines are compared to solve the \emph{same}
static coalition problem, formulated based on one snapshot of dynamic scene above.
\change{Although the end-to-end predictors by RL and HGAN
do not require communication,
the overall performance is much lower than KS4 and NAC (147, 178 vs. 185.2, 187).}
All baselines are given the same number of computation and communication rounds.
Evolution of the total utility along with the iterations is shown in Fig.~\ref{fig:static_utiltiy}.
Our method \change{NAC} has a faster convergence than both \change{KS2}, \change{KS3} and \change{KS4}
($0.5s$ vs. $1s$, $3s$ and \change{$20s$}),
but with the highest utility after convergence ($187$ vs $181$, $185$ and \change{$185.2$}).
It confirms our analyses that larger $\mathcal{K}$ requires longer sequence of transformation
and thus slower to converge.
The proposed \change{NAC} predicts~$\{k_i\}$ non-uniformly and accelerates
the convergence even further with a good initial solution.
\change{Note that
since {KS4} can not converge at each iteration under limited computation time,
it achieves an utility less than NAC.}
\change{Both {GreedyNE} and {FastMaxSum} reach convergence within only $0.1s$ and $0.2s$,
however with a final utility significantly less than {KS4} and {NAC}
($150$,~$147$ vs. {$185.2$},~$187$).}
Lastly,
compared with \change{NAC}, \change{NACK} converges slower (\textbf{$2s$})
with a worse utility ({$186$}).
\change{Although trained by data from {KS3},
{NACInit} outperforms \change{KS3}
by a higher initial and final utility ({$185$ and $187$})}.
This verifies that both the parameter~$\mathcal{K}$ and the initial solution~$\nu_0$
are worth learning to accelerate the coordination.

\begin{figure}[!t]
  \centering
  \includegraphics[width=0.75\hsize]{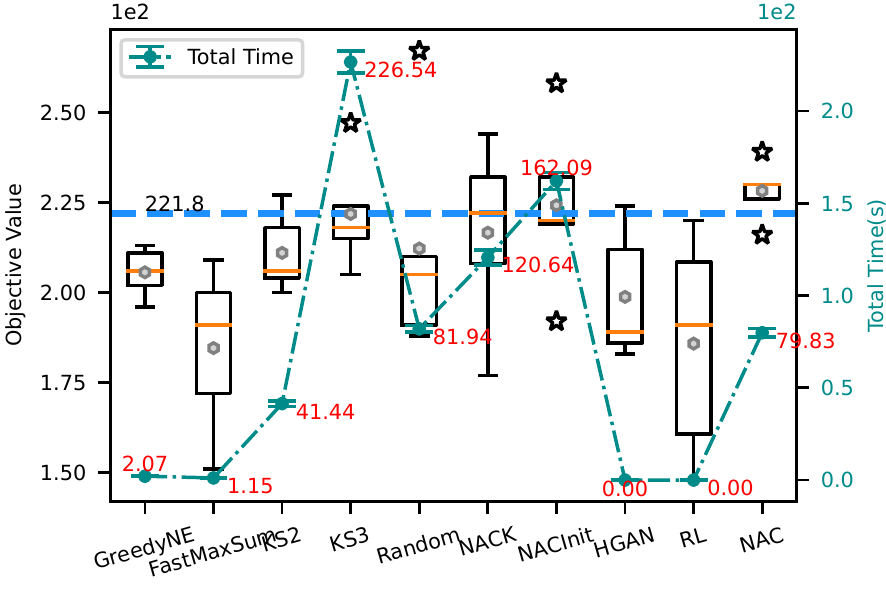}
  \vspace{-4mm}
  \caption{\change{Overall utility and total time elapsed in dynamic scenes.}
}
  \vspace{-2mm}
  \label{fig:dynamic_result}
\end{figure}
\begin{figure}[t!]
  \begin{subfigure}
  \centering
  \includegraphics[width=0.64\linewidth]{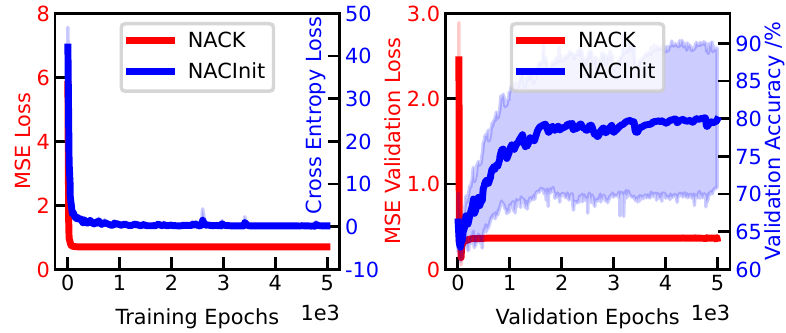}
  \end{subfigure}
  \begin{subfigure}
    \centering
    \includegraphics[width=0.34\linewidth]{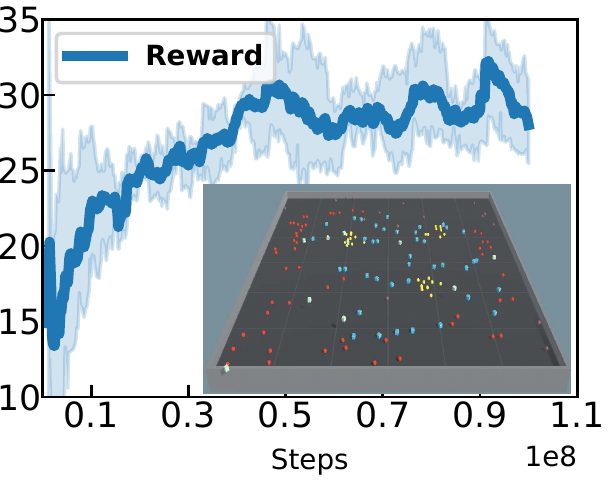}
  \end{subfigure}
    \vspace{-5mm}
    \caption{Training loss (\textbf{left})
      and validation loss (\textbf{middle}) for HGAN-K and HGAN-Init;
    \change{training reward for method RL~\cite{de2021decentralized} (\textbf{right}).}}
\label{fig:gnn_learning}
\vspace{-4mm}
\end{figure}


\subsubsection{Communication Overhead}
All methods are executed for~$20$ iterations and the total amount of exchanged messages is compared.
As shown in Fig.~\ref{fig:static_utiltiy},
\change{it is clear that drastically more messages are generated as the parameter~$\mathcal{K}$ increases
i.e., $50$, $680$, $9094$ and $113496$ for GreedyNE, KS2, KS3 and KS4}, respectively.
Both \change{NACK} and \change{NACInit} generate less messages than \change{KS4}
($6443$ and $8064$ vs. \change{$113496$}),
due to either smaller~$\mathcal{K}$ or faster convergence.
Notably, our \change{NAC} requires merely~$3116$ messages to reach convergence,
drastically less than both variants \change{NACK} and \change{NACInit}.
This signifies the effect of acceleration via the \change{NAC}.
\change{Lastly, {FastMaxSum} requires more messages than {KS2}
but less than {KS4} however with different structure.}
\change{Note that although KS4 yields a higher utility,
it takes\textbf{~$6$ times} more time and\textbf{~$12$ times} more messages compared with KS3.
Thus, KS4 is unsuitable for real-time applications in dynamic scenarios as discussed below.}

\subsubsection{Long-term Performance over Dynamic Scenes}
The described experiment is repeated~$10$ times
with different random seeds for~$200$ time steps.
As summarized in Fig.~\ref{fig:dynamic_result},
\change{our~\change{NAC} drastically improves the accumulated performance by~$11\%$
in average compared with {GreedyNE}, $24\%$ than {FastMaxsum},
$15\%$ than HGAN and $23\%$ than RL}.
\change{
Moreover, NAC exhibits a smaller variance in the overall utility thus generating more
stable solutions than both end-to-end predictors RL and HGAN}.
In contrast, \change{KS3} without the NAC achieves an overall
utility~$8\%$ higher than GreedyNE.
Similar to the static scenario, both~\change{NACK} and~\change{NACInit} have
worse performance compared with~\change{NAC} ($216$, $224$ vs. $228$).
\change{This indicates that both branches of the proposed~{NAC},
i.e., the prediction of both~$\nu_0$ and $\mathcal{K}$, can
facilitate faster adaptation in dynamic scenarios thus improving long-term performances.}

\subsection{Generalization Analyses}\label{subsec:general}
To verify how the proposed~NAC generalizes to different scenes and system sizes,
the following two aspects are tested.

\subsubsection{Validation Accuracy}
As shown in Fig.~\ref{fig:gnn_learning},
both the {HGAN-K} and {HGAN-Init} \change{(similar to~HGAN
  from \cite{zhou2022graph})} models converge after
respectively $500$, $2000$ epochs during learning.
Both models are tested in different validation sets,
by running diverse scenarios of different random seeds.
It can be seen that the HGAN-K model has close to~$0.2$ MES loss,
while the HGAN-Init model reaches~$80\%$ accuracy.
\change{
In comparison, the RL method~\cite{de2021decentralized} takes much more iterations
  to train with a much slower convergence.}

\subsubsection{Novel Scenarios}
\change{
  As summarized in Table~\ref{tab:result}, the proposed NAC and other baselines
  are evaluated in \textbf{four} different scenarios with
various system sizes and initial states in 200 time steps.
It can be seen that our \change{NAC} outperforms all baselines in all four cases in terms of mean utility 
over 10 tests with different random seeds,
indicating that the proposed scheme is robust
against variations of system sizes,
particularly the ratio between swat and target robots.
In addition,
the end-to-end predictors RL and HGAN exhibit
a large variance in performance across different scenarios in Table~\ref{tab:result},
meaning that the end-to-end predictors often generalize poorly to novel scenarios
that are different from the training set.
}

\begin{table}[t!]
  \centering
  \scalebox{1.0}{
    \begin{tabular}{ccccc}
    \toprule
    \multicolumn{1}{c}{\multirow{4}{*}{\textbf{Method}}} & \multicolumn{4}{c}{\textbf{Overall Utility}}\\
    \cmidrule(lr){2-5}
     & \begin{tabular}[c]{@{}c@{}}\textbf{Case1}\\ swat:40\\ target:20\end{tabular} & \begin{tabular}[c]{@{}c@{}}\textbf{Case2}\\ swat:40\\ target:100\end{tabular} & \begin{tabular}[c]{@{}c@{}}\textbf{Case3}\\ swat:150\\ target:100\end{tabular} & \begin{tabular}[c]{@{}c@{}}\textbf{Case4}\\ swat:150\\ target:20\end{tabular} \\
     \midrule
    GreedyNE & $62 \pm 19$ & $296 \pm 38$ & $1161 \pm 76$ & $198 \pm 37$ \\
    FastMaxSum & $31 \pm 14$ & $222 \pm 33$ & $922 \pm 93$ & $195 \pm 26$ \\
    \change{HGAN} &$\change{63} \pm \change{14}$ &$\change{282} \pm \change{31}$ &$\change{990} \pm \change{88}$ &$\change{206} \pm \change{27}$ \\
    \change{RL} &$\change{70} \pm \change{31}$ &$\change{256} \pm \change{75}$ &$\change{972} \pm \change{100}$ &$\change{196} \pm \change{57}$ \\
    \textbf{NAC} & {$\mathbf{74} \pm \mathbf{13}$} & {$\mathbf{316} \pm \mathbf{10}$} & {$\mathbf{1202} \pm \mathbf{50}$} & {$\mathbf{211} \pm \mathbf{17}$} \\
    \bottomrule
    \end{tabular}
  }
  \vspace{2mm}
  \caption{\change{Comparisons in novel scenarios.}}
  \label{tab:result}
  \vspace{-8mm}
\end{table}

\begin{figure*}[!htb]
  \centering
  \includegraphics[width=0.78\linewidth]{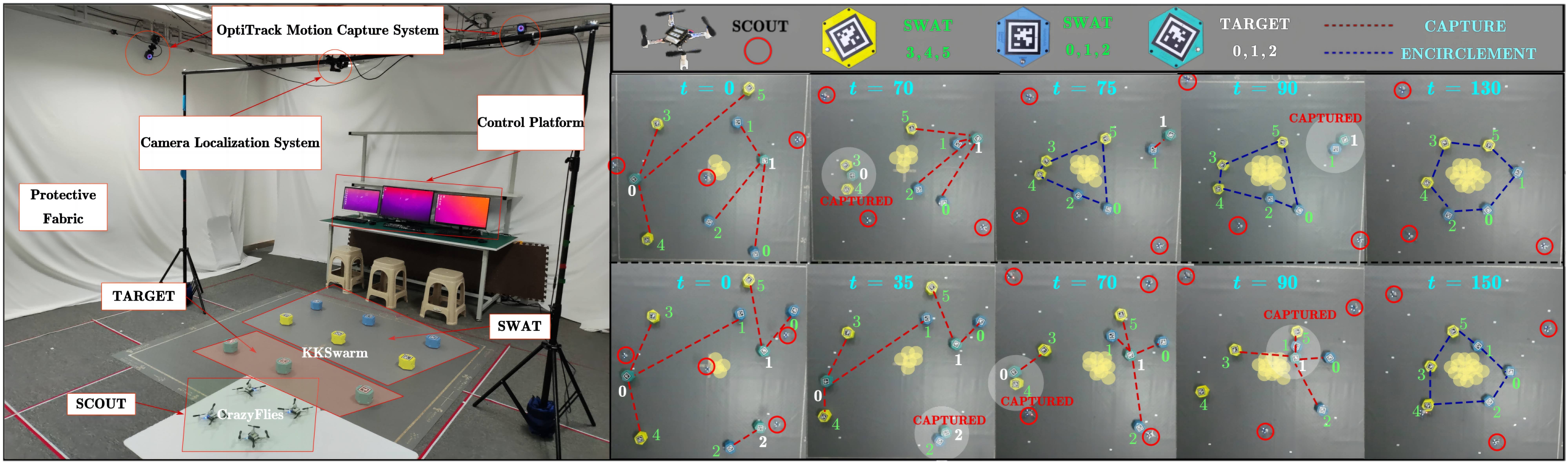}
  \vspace{-2mm}
  \caption{
      Snapshots of the hardware experiments:
      experimental setup (\textbf{left}),
      $11$ robots (\textbf{right-top})
      and $13$ robots (\textbf{right-bottom}),
      where UAVs are SCOUT robots and UGVs are TARGET and SWAT robots.
    }
  \vspace{-3mm}
  \label{fig:real-exp}
\end{figure*}
\vspace{-2mm}
\subsection{Hardware Experiment}
\change{Hardware experiments are conducted on a large fleet of UAVs and UGVs,
close to the maximum capability of our lab}.
Detailed descriptions of the setup including state estimation,
motion control and software architecture can be found in the supplementary material.
\change{Due to the modularization of our system,
most existing localization techniques can be integrated to provide state estimation,
such as GPS, RTK, UWB and SLAM-based localization.}
Fig.~\ref{fig:real-exp} contains the snapshots
where the proposed NAC is applied to two distinct scenarios,
i.e., $11$~robots ($6$~SWAT, $2$~TARGET, $3$~SCOUT) and $13$~robots
($6$~SWAT, $3$~TARGET, $4$~SCOUT)
with different initial states and distribution of resources.

More specifically,
in the first scenario shown in Fig.~\ref{fig:real-exp},~$3$ SCOUT robots periodically
are assigned to perform the exploration tasks in these regions with higher uncertainty.
Initially SWAT robots $3,4$ are assigned to TARGET robot $0$ while other SWAT
robots move towards TARGET robot $1$.
At $t = 70s$, TARGET robot $0$ is captured by the proposed collaborative strategy.
Afterwards,
the SWAT robots $3,4$ switch to the capture task of TARGET robot~$1$,
while other SWAT robots encircle the resources in the middle.
This is because the utility of performing the encirclement task is higher than the capture task of TARGET robot~$1$.
Once TARGET robot~$1$ is finally captured at~$t = 90s$,
all SWAT robots return the encirclement task but with a different topology, until the termination at $t=130s$.
In the second scenario shown in Fig.~\ref{fig:real-exp},
more TARGET robots and SCOUT robots are deployed in differently initial positions.
It results in a~$98\%$ coverage in~$100s$,
compared with only~$90\%$ in~$130s$ for the first scenario.
An immediate outcome is that switches of task assignment
  happen more frequently,
during which all tasks are finished at~$t = 150s$.
\change{These results demonstrate that the proposed method can be deployed
to rather large-scale robotic fleets.}
Experiment videos can be found in the supplementary material.

\section{Conclusion \& Future Work} \label{sec:conclusion}
A fully-distributed and generic coalition strategy called KS-COAL
is proposed in this work.
Different from most centralized approaches,
it requires only local communication
and ensures a K-serial Stable solution upon termination.
Furthermore, HGAN-based heuristics are learned to accelerate KS-COAL
during dynamic scenes by selecting more appropriate parameters and promising initial solutions.
\change{
  Practical concerns such as communication delays, motion and perception uncerainties
  might degrade the effectiveness of the method,
  which is part of our ongoing investigation.}
{
  Future work involves (i) the consideration of temporal constraints between collaborative tasks
  such as precedence and concurrency,
  which requires more long-term planning;
  (ii) the learning and prediction of adversarial behaviors for large-scale games.}


\bibliographystyle{IEEEtran}
\bibliography{contents/references}

\end{document}